%% file: paper.tex
\documentclass[10pt,twocolumn,letterpaper]{article}

\usepackage{iccv}
\usepackage{times}
\usepackage{epsfig}
\usepackage{graphicx}
\usepackage{amsmath}
\usepackage{amssymb}
\usepackage{caption}
\usepackage{bm}

\usepackage[normalem]{ulem}

\renewcommand{\vec}[1]{\bm{#1}}
\newcommand{\myb}{\vec{\beta}}
\newcommand{\myt}{\vec{\theta}}
\newcommand{\myg}{\vec\gamma}

\usepackage[pagebackref=true,breaklinks=true,letterpaper=true,colorlinks,bookmarks=false]{hyperref}

\iccvfinalcopy 

\def\iccvPaperID{6034} 

\ificcvfinal\fi
\begin{document}

\title{Three-D Safari: Learning to Estimate Zebra Pose, Shape, and
  Texture\\ from Images ``In the Wild''}

\author{Silvia Zuffi$^1$\hspace{0.05\linewidth} Angjoo Kanazawa$^2$
\hspace{0.05\linewidth} Tanya Berger-Wolf$^3$\hspace{0.05\linewidth} Michael J. Black$^4$\\
$^1$IMATI-CNR, Milan, Italy, 
$^2$University of California, Berkeley\\
$^3$University of Illinois at Chicago\\
$^4$Max Planck Institute for Intelligent Systems, T\"{u}bingen, Germany\\
{\tt\small silvia@mi.imati.cnr.it},
{\tt\small kanazawa@berkeley.edu},
{\tt\small tanyabw@uic.edu},
{\tt\small black@tuebingen.mpg.de}}

\twocolumn[{%
\renewcommand\twocolumn[1][]{#1}%
\maketitle
\begin{center}
    \newcommand{\teaserwidth}{\textwidth}
\vspace{-0.3in}
    \centerline{
   \includegraphics[width=\teaserwidth,clip]{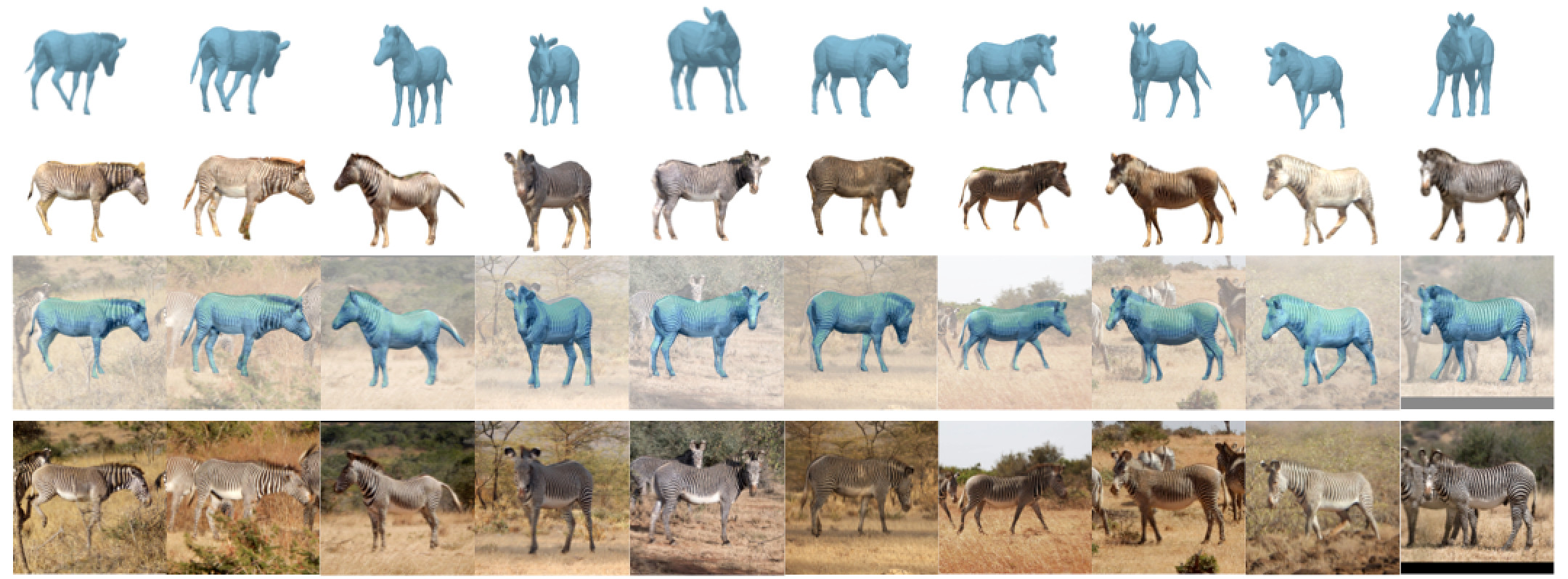}
     }
\vspace{-0.1in}
   \captionof{figure}{{\bf Zebras from images}. 
We automatically extract 3D textured models of zebras from in-the-wild
images. We regress directly from pixels, without keypoint detection or segmentation.
}
\label{fig:teaser}
\end{center}%
}]

\maketitle

\input{source/abstract}


\input{source/introduction}

\input{source/previous}

\input{source/approach}

\input{source/method}

\input{source/experiments}

\input{source/conclusion}

{\small
\bibliographystyle{ieee_fullname}
\bibliography{refs}
}

\end{document}

%% file: source/abstract.tex
\begin{abstract}
We present the first method to perform automatic 3D pose, shape and texture capture of animals from images acquired in-the-wild. In particular, we focus on the problem of capturing 3D information about Grevy's zebras from a collection of images. The Grevy's zebra is one of the most endangered species in Africa, with only a few thousand individuals left. Capturing the shape and pose of these animals can provide biologists and conservationists with information about animal health and behavior. In contrast to research on human pose, shape and texture estimation, training data for endangered species is limited, the animals are in complex natural scenes with occlusion, they are naturally camouflaged, travel in herds, and look similar to each other.  To overcome these challenges, we integrate the recent SMAL animal model into a network-based regression pipeline, which we train end-to-end on synthetically generated images with pose, shape, and background variation. Going beyond state-of-the-art methods for human shape and pose estimation, our method learns a shape space for zebras during training. 
Learning such a shape space from images using only a photometric loss is novel, and the approach can be used to learn shape in other settings with limited 3D supervision.
Moreover, we couple 3D pose and shape prediction with the task of texture synthesis, obtaining a full texture map of the animal from a single image. 
We show that the predicted texture map allows a novel per-instance unsupervised optimization over the network features.
This method, SMALST (SMAL with learned Shape and Texture) goes beyond previous work, which assumed manual keypoints and/or segmentation, to regress directly from pixels to 3D animal shape, pose and texture.
Code and data are available at \url{https://github.com/silviazuffi/smalst}.
\end{abstract}

%% file: source/introduction.tex
\section{Introduction}
Rapid progress has been made on estimating 3D human pose, shape, and
texture from images.  
Humans are special and, consequently,
the amount of effort
devoted to this problem is significant.  
We scan and model the body, label images by hand, and build motion
capture systems of all kinds.
This level of investment is not possible for every animal species.
There are simply too many and the scientific community interested in a particular
species may not have the resources for such an effort.
This is particularly true of endangered species.
We focus on one of the most endangered animal
species, the Grevy's zebra, for which only about 3000 individuals
remain~\cite{Rubenstein2018}.
Here we describe a new deep-learning method to regress 3D animal shape, pose, and
texture directly from image pixels (Figure \ref{fig:teaser}) that does not require extensive image
annotation, addresses the key challenges of animals in the wild, and
can scale to large data collections.
This provides a new method that can be extended to other species (see Sup.~Mat.~for results on horses).

In general, there is very little previous work on estimating animal
shape and pose. Existing methods require manual annotation
of the test images with keypoints and/or segmentation
\cite{Zuffi:CVPR:2018,Zuffi:CVPR:2017} or clean imaging conditions
where automatic segmentation is possible  \cite{biggs2018creatures}.

\begin{figure}[t]
    \centerline{
   \includegraphics[width=0.55\columnwidth]{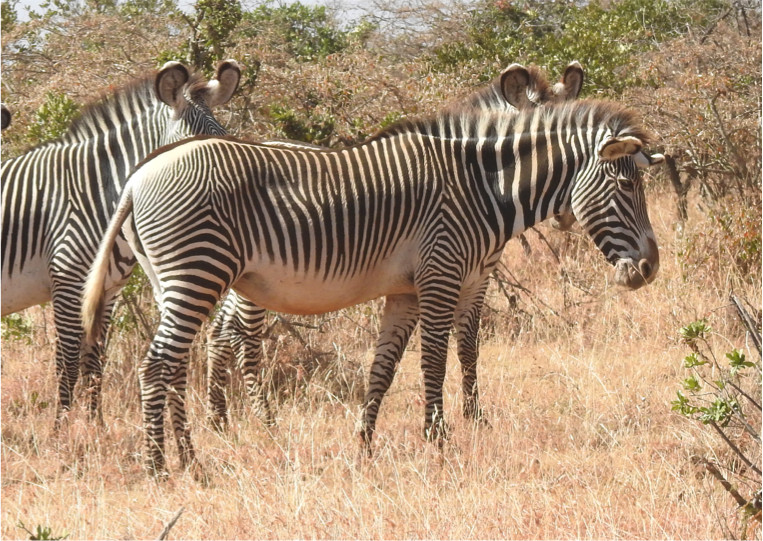}
   \includegraphics[width=0.45\columnwidth]{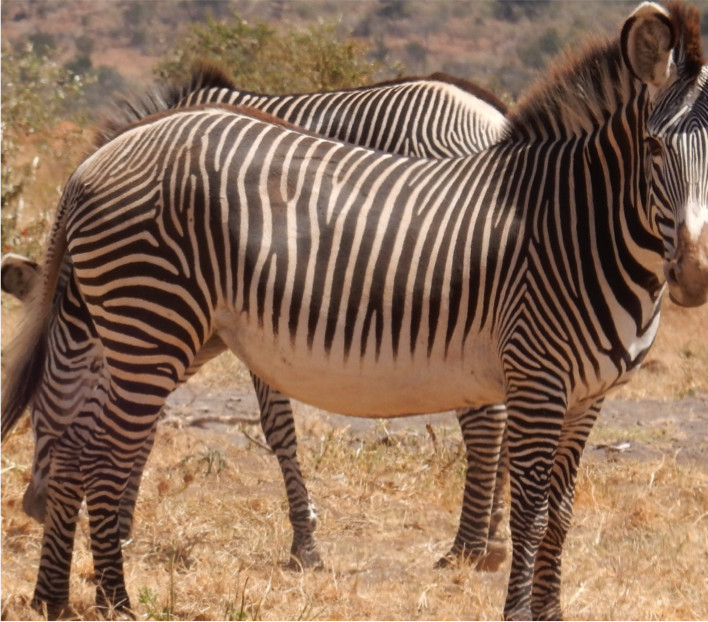}
     }
   \captionof{figure}{{\bf Challenges of animals in the wild}. Challenging images
     for keypoint detection and assignment (left), where the zebra in the foreground
   appears to have three front legs; and  for segmentation (right),
   where the back legs of the foreground animal are hard to distinguish from the neck of the zebra in the background.
   } 
\label{fig:issues}
\end{figure}

Animals present unique challenges relative to humans  (see Figure \ref{fig:issues}). 
First, animals often live in environments where
their appearance is camouflaged, making bottom-up approaches such as automatic
segmentation a challenge. 
Second, animals like zebras live in herds, where
overlapping subjects of similar appearance make reliable keypoint extraction
challenging.
Third, in comparison to the
study of human body pose and shape, the amount of data is limited,
particularly for endangered animals, where 3D scanning is infeasible.
Thus, although humans and animals are both deformable articulated
objects, the lack of training data makes naive application of current deep learning
methods that work for humans 
impractical for animals.

\begin{figure*}
    \centerline{
   \includegraphics[width=0.25\columnwidth]{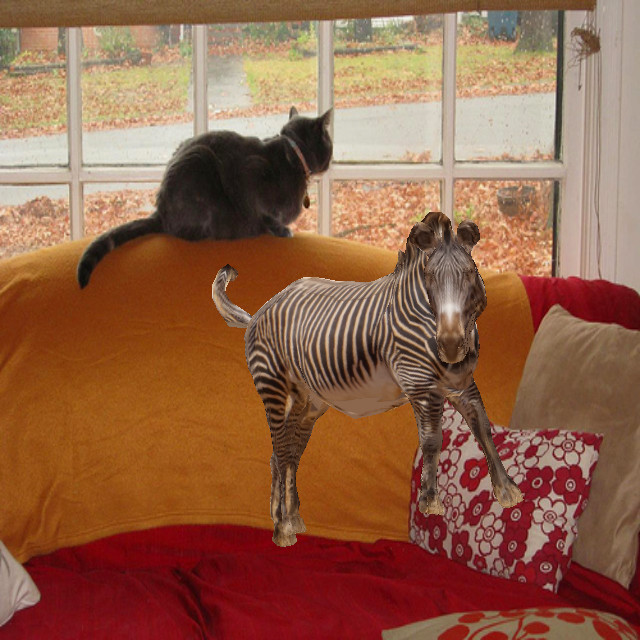} \includegraphics[width=0.25\columnwidth]{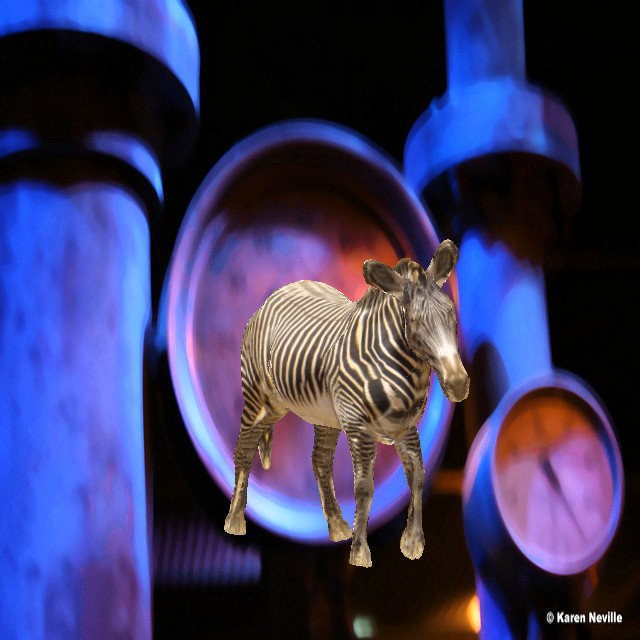}
   \includegraphics[width=0.25\columnwidth]{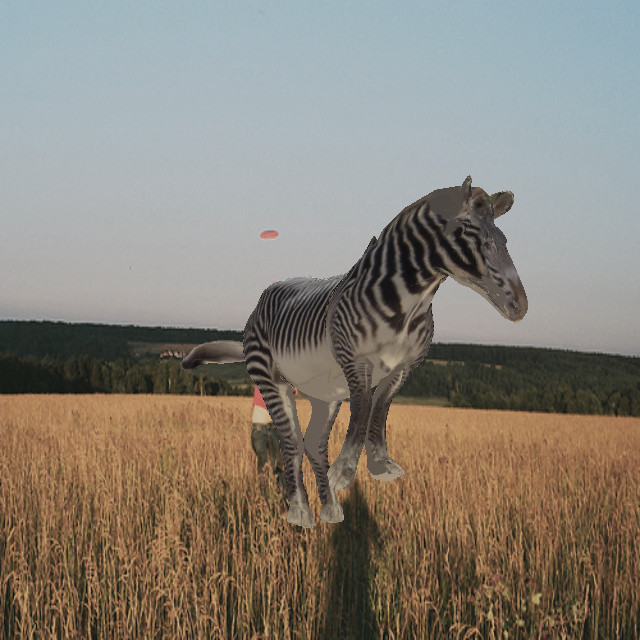} \includegraphics[width=0.25\columnwidth]{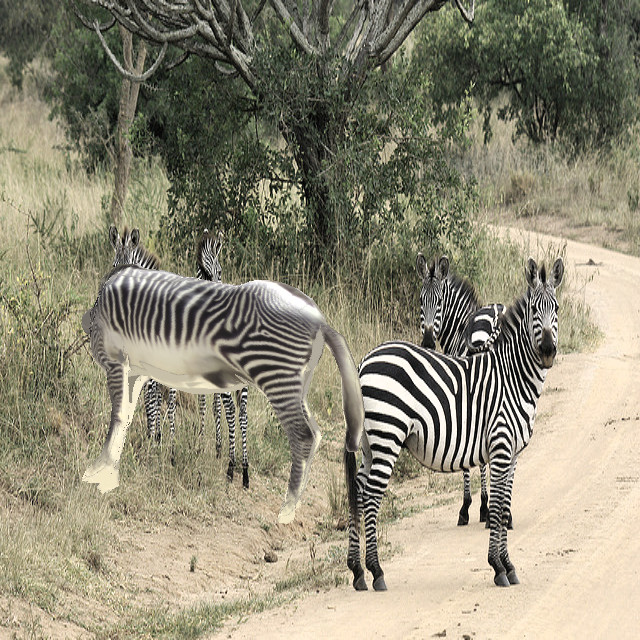}
   \includegraphics[width=0.25\columnwidth]{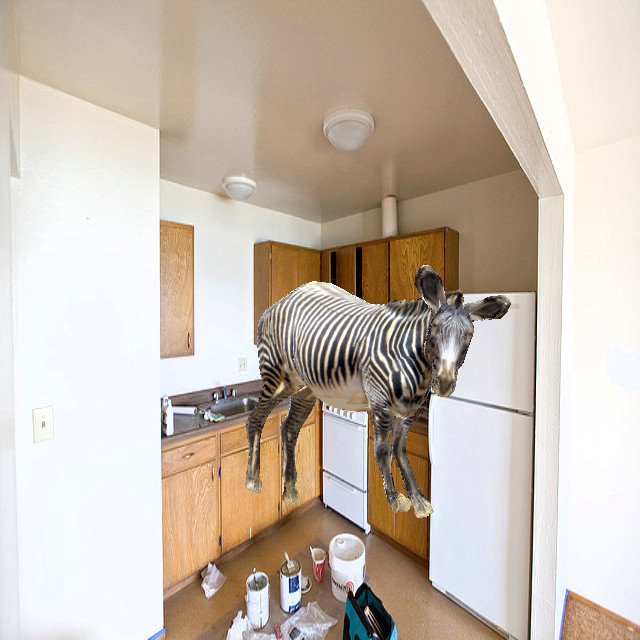} \includegraphics[width=0.25\columnwidth]{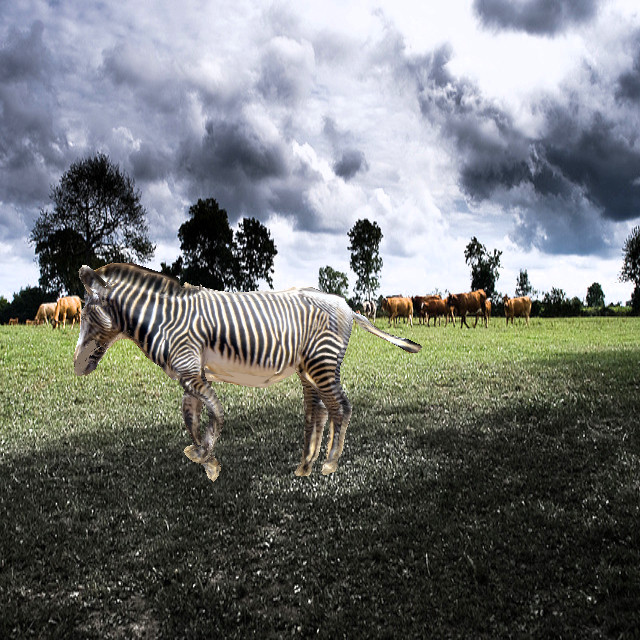}
   \includegraphics[width=0.25\columnwidth]{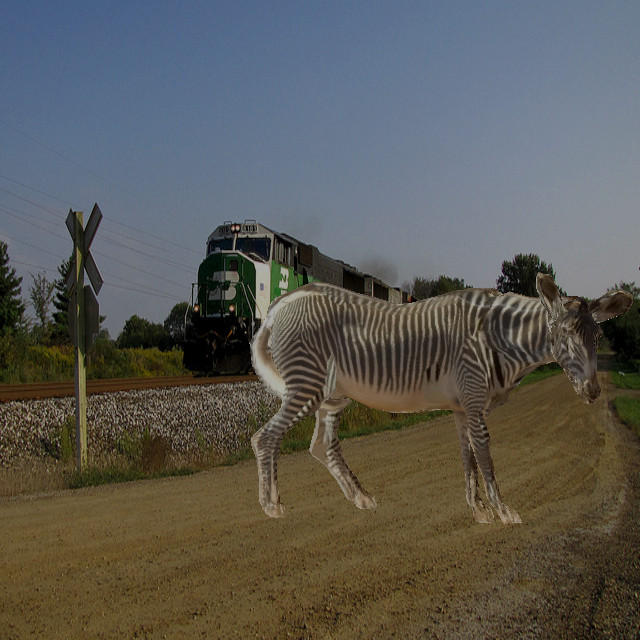} \includegraphics[width=0.25\columnwidth]{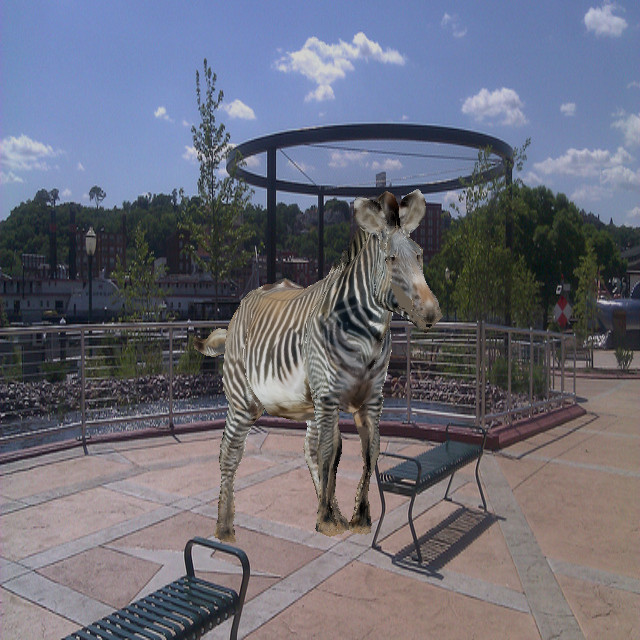}
     }
    \centerline{
   \includegraphics[width=0.25\columnwidth]{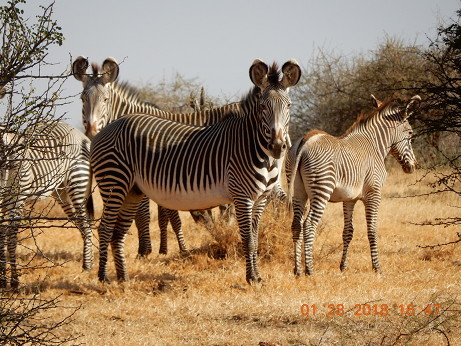} \includegraphics[width=0.25\columnwidth]{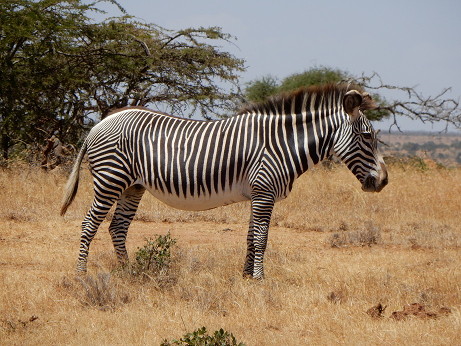}
   \includegraphics[width=0.25\columnwidth]{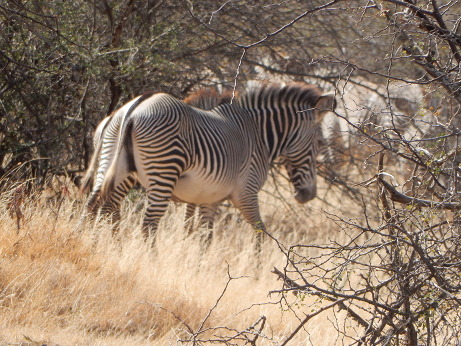} \includegraphics[width=0.25\columnwidth]{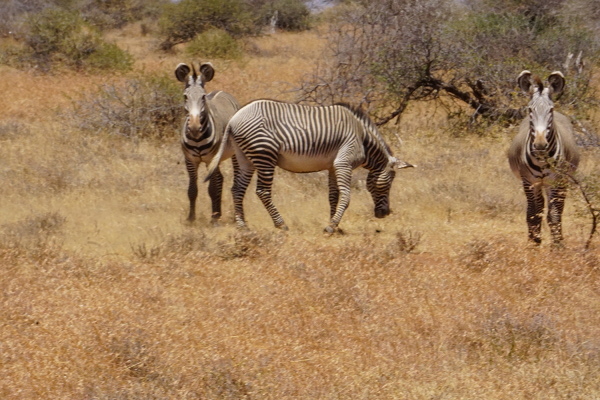}
   \includegraphics[width=0.25\columnwidth]{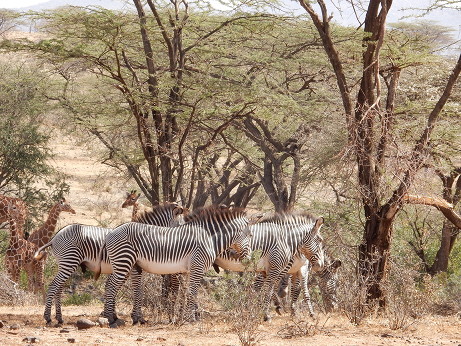} \includegraphics[width=0.25\columnwidth]{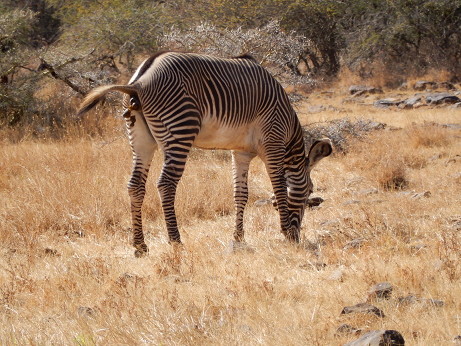}
   \includegraphics[width=0.25\columnwidth]{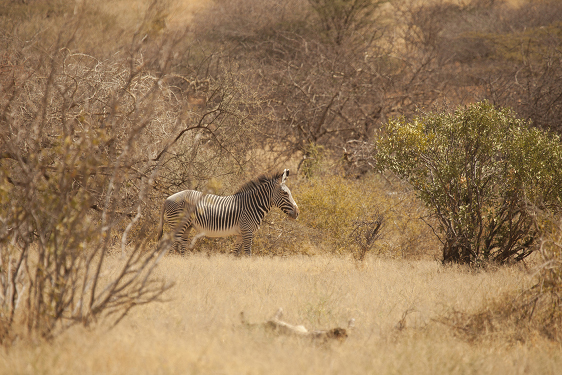} \includegraphics[width=0.25\columnwidth]{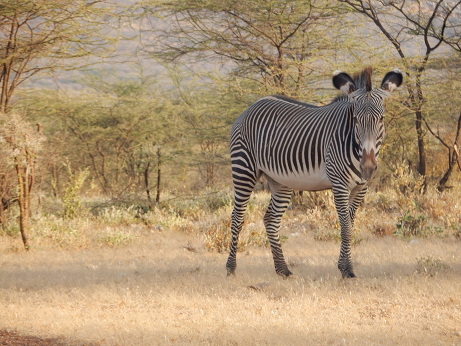}
     }
   \captionof{figure}{{\bf Datasets}. Example of images from the digital dataset (top) and real dataset (bottom).
   } 
\label{fig:datasets}
\end{figure*}

We overcome the lack of data by exploiting synthetic data and an image
reconstruction loss using a generative,  analysis-by-synthesis,
approach.
While accurate synthetic human models exist for training, animals
models of sufficient quality are rare, particularly for endangered  species.
A novelty of our approach is that instead of using completely synthetic data, we
capture the texture of the animals from real images and render them with
variability of background, pose, illumination and camera.  This is obtained
exploiting the recent SMALR method \cite{Zuffi:CVPR:2018}, which allows us to obtain accurate
shape, pose, and texture of 10 animals by annotating only about 50 images.
From this, adding variations to the subjects, we generate thousands of synthetic training images (Figure \ref{fig:datasets}).
We demonstrate that these are realistic enough for our method to learn
to estimate body shape, pose and texture from image pixels without any 
fine-tuning on additional hand-labeled images.

We go beyond previous works in several important ways and our main contributions are as follows.
Using a fully generative model of animal shape, appearance, and
neural rendering, we use a photometric loss to train 
a neural network that predicts the 3D pose, shape, and texture map of
an animal from a single image.
A key novelty of the network is that it links the texture prediction to 3D pose and shape through a shared feature space,
such that, in predicting the texture map, the network estimates model parameters
for an optimal mapping between image pixels and texture map elements. 
In order to prevent the network from just learning average texture map
colors, inspired by \cite{cmrKanazawa18}, we predict the  flow between the image pixels and the texture map.
We go beyond \cite{cmrKanazawa18},  however, to deal with an
articulated object with a much more complex texture map containing multiple,
disconnected regions.

We base our method on a recently introduced 3D articulated shape model of animals, the SMAL model~\cite{Zuffi:CVPR:2017},
which can represent animals of different shape with a low-dimensional linear model learned from a small set of toys.
Instead of relying on the SMAL model shape space,  we compute shape variations with a network layer. 
This corresponds to learning
a novel shape space during training and predicting the shape coefficients in this space at test time.
We have not seen this sort of neural shape learning before and this is key for endangered species where it would be difficult to build
an accurate a priori shape model.
Additionally, unlike most human pose estimation methods, we estimate
camera focal length.  This is important for animals where the elongated shape can
result in significant foreshortening.
Finally, we also show that, since our network predicts a full texture map, we can
exploit the photometric loss to perform a per-instance optimization of the
model parameters at test time by searching in the network feature space. By
using a background model we are able to refine and obtain more detailed pose and
shapes in a fully automatic manner without relying on any segmentation masks at test time.
Figure \ref{fig:network} provides an overview of the approach.
We call our method SMALST, for {\em SMAL with learned
  Shape and Texture.}
We test the approach on a dataset of 200 individual zebra images where we evaluate
the method both quantitatively and qualitatively.

%% file: source/previous.tex
\section{Previous work}

{\bf Human 3D Pose and Shape Estimation.}
The amount of work in the field of 3D human pose estimation is huge; here we review monocular model-based approaches that 
estimate shape and pose, as they are most related to the goal of this
paper.
The estimation of 3D pose and shape from monocular images is largely based on low dimensional statistical models of the human body learned from thousands of scans~\cite{Allen:2003, Anguelov05,SMPL:2015}.
Currently the most popular model is SMPL~\cite{SMPL:2015}
and recent work on modeling animals also builds on its formulation.
Low-dimensional parametric models are particularly attractive for neural architectures as they can generate high quality meshes from a small number of network-generated parameters.
Tan \etal~\cite{TanBC17} train a network that learns the mapping from silhouettes to pose and shape parameters of the SMPL model.
Omran \etal~\cite{omran2018NBF} exploit SMPL in conjunction with a bottom-up  body part segmentation.
Pavlakos \etal \cite{PavlakosCVPR2018} estimate 3D body pose and shape of the SMPL model using a two-stage architecture
based on 2D keypoints and silhouettes.

Methods focused on humans have the advantage of large datasets of images with ground truth 3D poses captured indoor with mocap systems \cite{Ionescu2014}.
We do not have this for animals and indoor imagery does not generalize
well to the wild.
One way for obtaining approximate 3D pose and shape from outdoor images is to use human annotators. 
Lassner \etal \cite{Lassner:UP:2017} build a dataset of human-rated high-quality 3D model fits.
von Marcard \etal \cite{vip:eccv:2018} introduce an in-the-wild
dataset of human 3D pose and shape obtained by exploiting IMU sensors
and video.
Alternatively, Varol \etal~\cite{VarolCVPR2017}  create SURREAL, a full synthetic dataset of 3D pose and shape.
 Tung \etal \cite{tung2017self} exploit temporal consistency over video frames to
train an end-to-end prediction model on images without ground truth 3D pose and
shape.  Kanazawa \etal \cite{hmrKanazawa17} exploit adversarial
training, which uses decoupled datasets of 2D keypoints and 3D pose and
shape to train on in-the-wild images in a weakly-supervised manner.

Model-based methods that capture texture are becoming popular with the goal of creating rigged human avatars.
Bogo \etal~\cite{Bogo:ICCV:2015} create 3D models of shape and appearance from RGB-D sequences.
Alldieck \etal~\cite{alldieck2018video} go further to create 3D
textured models with approximate clothing shape from multiple video frames of a subject in a reference pose.

Deep learning methods for human pose and shape estimation  combine powerful 2D joint detectors, 
accurate body part segmentations, large datasets of
human 3D pose, and an expressive articulated 3D shape model.
None of the previous work exploits appearance, likely because it
varies so much due to clothing.
For birds, not humans, Kanazawa \etal \cite{cmrKanazawa18} learn a generative model of appearance by
modeling the way surface colors are mapped from the image to a texture map. 
In our work we explore this approach to appearance modeling in conjunction with a SMPL-style model of animals with the intuition that learning to predict texture helps in the task of pose and shape recovery.

{\bf Animals Pose and Shape Estimation.}
Animals are presented in many object-recognition datasets where methods for bounding box detection, recognition, and
instance segmentation are common \cite{he2017maskrcnn}. 
There is very little work, however, on detecting 3D animal
shape and pose.
The seminal work on this from Cashman and Fitzgibbon
\cite{dolphins} shows how to estimate the shape of dolphins
from monocular images.
Ntouskos \etal \cite{Ntouskos2015} model articulated animals as a collection of 3D primitives
that they estimate from segmented images.
Vincente and Agapito \cite{Vicente2013} show how to build a rough giraffe shape
from two views.   
Kanazawa \etal~\cite{Kanazawa:Cats:2016} learn deformation models 
for cats and horses. Kanazawa \etal \cite{cmrKanazawa18} predict 3D shape and texture of birds from
images, without assuming a template model of the bird's 3D shape, but they do not
model pose.
Methods based on video have potentially more information about animal shape at their disposal.
Reinert \etal~\cite{Reinert:2016} show the extraction of
a rough animal shape from video in terms of generalized cylinders.
They also recover a texture map from one video frame.

In contrast to  humans, none of the previous approaches is based on 3D animal models
learned from scans, therefore they lack realism and fine detail.
Moreover,  previous work estimates the 3D shape of a single
animal; none address the problem of capturing shape for a 
large number of subjects of the same species with variable body shape.

Zuffi \etal \cite{Zuffi:CVPR:2017} introduced the SMAL model, a 3D
articulated shape model of animals, that can represent inter and intra
species shape variations.
They train the model from scans of toys, which may not exist for
endangered species or may not be accurate.
They go further in \cite{Zuffi:CVPR:2018} to fit the model to multiple
images, while allowing the shape to deform to fit to the individual shape of the animals. 
This allows them to capture shape outside of the SMAL shape space, increasing
realism and generalization to unseen animal shapes. 
Unfortunately, the method is
based on manually extracted silhouettes and keypoint annotations. 
More recently, Biggs \etal \cite{biggs2018creatures} fit the SMAL model to images
automatically by training a joint detector on synthetically generated silhouettes.
At inference time, their method requires accurate segmentation and is not robust to occlusion.

In biology, animal tracking is extremely important and many tools
exist.
Recently deep learning methods have been applied to help solving the
feature tracking problem.
Mathis \etal \cite{Mathis2018} use deep learning to track animal keypoints defined by a user on infrared images captured in the lab. 
They show applications to rodents, bees and other small animals.
So far, no methods address the shape and pose estimation problem we
tackle here.

%% file: source/approach.tex
\section{Approach}

We formulate the problem of estimating the 3D pose and shape of zebras from
a single image as a model-based regression problem, where we train a neural network to predict 3D
pose, shape and texture for the SMAL model.

An important aspect of our approach is that we rely on a digitally generated
dataset to train our network. While the quality of synthetic image generation has
  dramatically improved thanks to advancements in computer graphics rendering
  and 3D modeling approaches, such synthetic data are
  expensive to obtain and insufficient in terms of variation in shape and
  appearance. Unlike faces, where the community has developed realistic
  generative models of shape and appearance from a large amount of high quality
  3D scans, no convincing generative models of 3D animals exist.

Instead of relying on synthetic data that is not sufficiently realistic for representing animals,
we capture appearance from real images and use this data to render realistic samples with
pose, shape and background variation. This approach of mixing real and synthetic is not novel: Alhaija \etal \cite{Alhaija2017BMVC} use synthetic cars in a real scene;
here we do the opposite: use real subjects against random backgrounds. 
We use the SMALR method~\cite{Zuffi:CVPR:2018} to create instance-specific SMAL
models from images with a captured texture map (see an example in Figure \ref{fig:texmap} left).
We animate and render such models to create our digital training set. 

\subsection{SMAL model}

The SMAL model is a function $M(\myb, \myt, \myg)$ of shape
$\myb$, pose $\myt$ and translation $\myg$. $\myb$ is a vector of the coefficients
of the learned PCA shape space, 
$\myt \in \mathbb{R}^{3N} =\{\textbf{r}_i\}_{i=1}^{N}$ 
is the relative rotation, expressed with Rodrigues vectors, of the joints in the
kinematic tree, and $\myg$ is the global translation applied to the root
joint.
Unlike \cite{Zuffi:CVPR:2017}, which uses $N=33$ joints, we segment and add articulation to the ears, obtaining a model with $N=35$ body parts.
The SMAL function returns a 3D mesh, where the model template
is shaped by $\myb$, articulated by $\myt$ 
and shifted by $\myg$. 
Formally, let $\myb$ be a row vector of shape variables, then vertices of
a subject-specific shape in the reference T-pose are computed as:
\begin{equation}
\textbf{v}_{shape}(\myb) = \textbf{v}_{horse} + B_s \myb,
\label{eq:smal}
\end{equation}
 where $\textbf{v}_{horse}$ represents the vertices
of the SMAL model template and $B_s$ is a matrix of
deformation vectors.
In this work we focus on an animal in the Equine family, thus the template $\textbf{v}_{horse}$ is 
the shape that corresponds to the mean horse in the original SMAL model.
Given a set of pose variables $\myt$ and
global translation $\myg$, the model generates 3D mesh vertices
$\textbf{v}=M(\myb, \myt, \myg)$ 
in the desired pose by articulating $\textbf{v}_{shape}$ with linear blend skinning.
 
\textbf{SMALR}. Zuffi \etal \cite{Zuffi:CVPR:2018} show that aligning the SMAL model to images
and then allowing the model vertices to deviate from the  linear shape model can create realistic textured 3D models of unseen animals from images. 
The method is based on an initial stage where SMAL is fitted to a set of images of the same animal from 
different views and poses, obtaining an estimation of global shape parameters $\hat{\myb}$, 
per image translation $\hat{\myg}_i$, pose $\hat{\myt}_i$, and camera focal length $\hat{f}_i$.
In a refinement step, shape parameters, translations, poses and cameras are held fixed, but the model shape is allowed to change
through a set of displacement vectors:
\begin{equation}
\textbf{v}_{shape}(\textbf{dv}^{\textrm{SMALR}}) = \textbf{v}_{horse} +
 B_s \hat{\myb} + \textbf{dv}^{\textrm{SMALR}},
\label{eq:smalr}
\end{equation}
where the superscript SMALR indicates that the displacement vector $\textbf{dv}$ is obtained with the SMALR method.

\subsection{Digital dataset}
We created a computer generated dataset of 12850 RGB images of single zebras. 
We applied SMALR to a set of 57 images of Grevy's zebras creating models for 10 different subjects.
For each zebra model we generated random images that differ in background, shape, pose, camera, and appearance.
Models are rendered with OpenDR \cite{loper14}.
Examples of the images are in Figure \ref{fig:datasets}, top. 

\textbf{Pose variation}. For each zebra model we generated 1000 images with different poses obtained by sampling 
a multivariate Gaussian distribution over the 3D Rodrigues vectors that describe pose.
The sampling distribution is learned from the 57 poses obtained with SMALR and a synthetic walking sequence.
We also add, for each zebra model, about 285 images obtained by adding noise to the 57 poses.
We also add noise to the reference animal translation, that we set at $\myg_0=[0.5, -0.1, 20]$. 
Changing depth varies the size of the animal as we use perspective projection.

\textbf{Appearance variation}. In order to vary the appearance of the zebras, we apply a white-balance algorithm to the textures, doubling the number of texture maps, while
on the generated images we randomly add noise to the brightness, hue and saturation levels. Moreover, we also randomly add lighting to the rendered scene.
We generate images with random backgrounds by sampling background images from the COCO dataset \cite{COCO}.

\textbf{Shape and size variation}. We increase the variability of the zebra shapes by adding noise to the shape variables (we use 20 shape variables).
In addition to size variations due to depth, we add size variation by adding noise to the reference camera, which has a reference focal length $f_0=4000$.

Images are created with size $(640,640)$. For each image we also compute the texture uv-flow that represents 
the mapping between image pixels and textels, and can be interpreted as the flow between the image and the texture map.
For each image we save the following annotation data: texture map $T_{gt}$, texture uv-flow $\textbf{uv}_{gt}$, silhouette $S_{gt}$, pose $\theta_{gt}$, 
global translation $\myg_{gt}$, shape variables $\myb_{gt}$, vertex displacements $\textbf{dv}_{gt}^{SMALR}$, landmark locations $K_{2D,gt}$. 
We use a total of 28 surface landmarks, placed at the joints, on the face, ears and tail tip.
These are defined only once on the 3D model template.

\subsection{Real datasets}

For evaluation, we have collected a new dataset of images of Gravy's zebra captured in Kenya with pre-computed bounding boxes.
Examples of the images are in Figure \ref{fig:datasets}, bottom.
We selected a set of 48 images of zebras that were not used to create the digital dataset, 
and we annotated them for 2D keypoints. We used this as validation set. 
We then selected 100 images as our test set, also avoiding zebras from the two sets above. 
For evaluation, we manually generated the segmentation mask for this set of images, and annotated the 2D keypoints. 
We mirror the images in order to double the test set data.

%% file: source/method.tex
\subsection{Method}

We design a network to regress a texture map $T$, $\textbf{dv}$, 3D pose $\theta$, translation $\gamma$ and focal length $f$ variables from an input image.
We implemented the SMAL model in PyTorch. 
The vertex displacements from the template are not computed as in SMAL (see Equation \ref{eq:smal}), but are estimated from the regression network.
Formally:
\begin{equation}
\textbf{v}_{shape}(\textbf{dv}) = \textbf{v}_{horse} + \textbf{dv}, 
\label{eq:pytsmal}
\end{equation}
where $\textbf{dv}$ is a displacement vector generated as output of a linear layer.
Consequently, given ground truth shapes from Equation \ref{eq:smalr}, we define ground truth vertex displacements for network training as: $\textbf{dv}_{gt}=B_s \myb_{gt}+\textbf{dv}_{gt}^{\textrm{SMALR}}$.
We use the Neural Mesh Renderer (NMR)~\cite{NMR} for rendering the model
and perspective projection.
\begin{figure*}
\centerline{
   \includegraphics[width=1.0\linewidth]{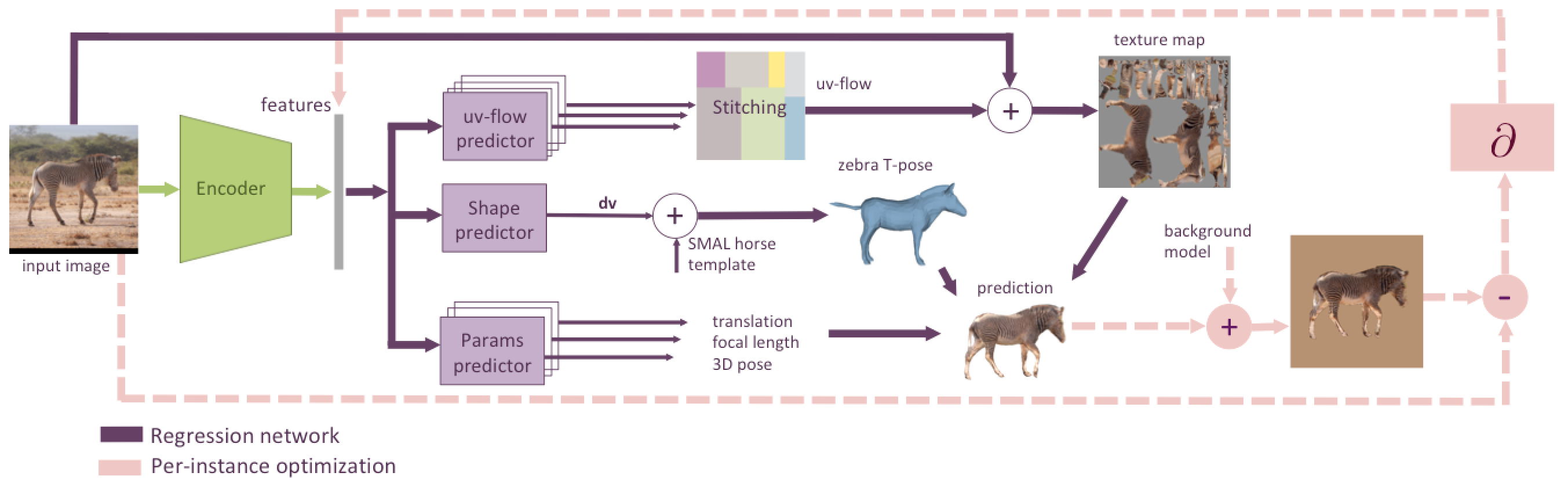}
}
\vspace{-0.1in}
   \caption{\textbf{Overall framework}. Given an input image the network predicts the uv-flow for each texture map sub-image (Fig. \ref{fig:texmap}) and then combines them to recover a full texture map.
   The vector displacements \textbf{dv} are added to the SMAL horse template to generate a 3D model in T-pose, which can be rendered given texture, pose, camera, and translation parameters.
   After prediction, we can perform per-instance optimization (dotted line), where we optimize over the feature space variables. }
\label{fig:network}
\end{figure*}

The regression network is illustrated in Figure \ref{fig:network}.
The encoder is composed of a Resnet 50 module that computes image features for an image of size $(256,256,3)$.
At training time the input image is a noisy bounding box computed given the ground truth segmentation mask. 
Ground truth values of translation and camera focus are modified accordingly when computing the bounding box.
At test time the input image is a pre-computed bounding box of the animal.
The Resnet module is followed by a convolutional layer with group normalization and leaky ReLU. 
The output of this layer is of size 2048.
We then add 2 fully connected layers, each of them composed by a linear layer, batch normalization and leaky ReLU.
We obtain as output a set of 1024 features.
From this set of features, we add independent layers that predict texture, shape, 3D pose, translation and camera focus.

\begin{figure}
    \centerline{
   \includegraphics[width=0.45\columnwidth]{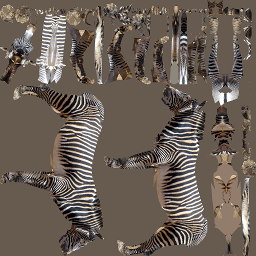}     \includegraphics[width=0.45\columnwidth]{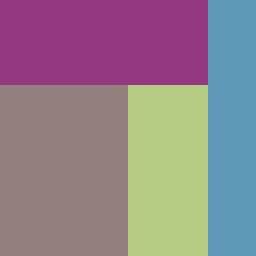} 
     }
    \captionof{figure}{{\bf Texture map}. Example of ground truth texture map (left) with sub-regions layout (right).
   } 
\label{fig:texmap}
\end{figure}

\textbf{Texture prediction}.
The texture prediction module is inspired by the work of Kanazawa \etal \cite{cmrKanazawa18}.
While \cite{cmrKanazawa18} explores texture
  regression on a simple texture map that corresponds to a sphere,
  quadrupeds, like zebras,
 have a  more complicated surface and texture map layout.
We therefore predict the texture map as a collection of 4 sub-images that we then stitch together.
We found this to work better than directly predicting the full texture map, probably because, given the complexity of the articulated model, 
the network has difficulty with the spatial discontinuities in the
texture map. 
We cut the texture map (that has size $(256,256)$) into 4 regions illustrated in Figure \ref{fig:texmap}.
For each sub-image we define an encoder and decoder.
Each encoder outputs a $(256,H,W)$ feature map, where H and W are a reduction of 32 of the size of the sub-image,
and is composed of 2 fully connected layers.
The decoders are composed of a set of convolutional layers and a final tanh module.
The output of the decoders is stitched to create a full uv-flow map,
that encodes which image pixels correspond to pixels in the texture map.

\textbf{Shape prediction}.
The shape prediction module is composed of a fully connected layer that outputs $40$ shape features $f_c$, and a linear layer
that predicts vertex deformations: 
\begin{equation}
\textbf{dv} = W f_s + b,
\label{eq:dv}
\end{equation}
where $b$ is a bias term.
$W$ is initialized with the SMAL blendshapes $B_s$ (see Equation \ref{eq:smal}).
We want to represent the shapes with a linear model that is more expressive than SMAL, therefore we seek to optimize the shape blendshapes through the network.
In order to limit the number of network parameters, we exploit the symmetry of the SMAL model and predict only half of the mesh vertices. 

\textbf{Pose prediction}.
The pose prediction module is a linear layer that outputs a vector of 3D poses as relative joint angles, expressed as Rodrigues vectors. 
The pose vector is of size $105$ as we use $35$ body joints.

\textbf{Translation prediction}. 
The translation prediction module is composed of two linear layers that predict translation in the camera frame and depth independently:
$\gamma_x=1.0+x$, $\gamma_y=y$, $\gamma_z=1.0+z+\gamma_{z,0}$, where $(x,y,z)$ are the outputs of the prediction layer, $\gamma_{z,0}=20$ as in the synthetic dataset. 
We add $1.0$ to the $x$ coordinate due to the distribution of the ground truth values in the training set.

\textbf{Camera prediction}. 
The camera prediction layer predicts the focal length of the perspective
camera. Since we also predict the depth in the network, 
 this parameter can be redundant; however we have found empirically that it allows better model fits to images.
The camera focal length is obtained as $f = f_0 + f_1 x$, where $x$ is the output of the prediction layer and $f_0=f_1=2700$.

\subsection{3D pose, shape and texture estimation}
We train the network to minimize the loss:
\begin{eqnarray}
L_{train} = L_{mask}(S_{gt}, S)+L_{kp_{2D}}(K_{2D,gt}, K_{2D})+ \nonumber \\
L_{cam}(f_{gt},f)+L_{img}(I_{input},I,S_{gt})+L_{pose}(\theta_{gt}, \theta)+ \nonumber \\
L_{trans}(\myg_{gt},\myg)+L_{shape}(\textbf{dv}_{gt},\textbf{dv})+ \nonumber \\
L_{uv}(\textbf{uv}_{gt}, \textbf{uv})+ L_{tex}(T_{gt},T)+L_{dt}(\textbf{uv}, S_{gt})
\end{eqnarray}
where: 
$S_{gt}$ is the mask, $L_{mask}$ is the mask loss, defined as the L1 loss between $S_{gt}$ and the predicted mask. 
$L_{kp_{2D}}$ is the 2D keypoint loss, defined as the MSE loss between $K_{2D,gt}$ and the projected 3D keypoints defined on the model vertices.
$L_{cam}$ is the camera loss, defined as the MSE loss between $f_{gt}$ and predicted focal length.
$L_{img}$ is the image loss, computed as the perceptual distance
\cite{zhang2018perceptual} between the masked input image and rendered zebra. 
$L_{pose}$ is the MSE loss between $\theta_{gt}$ and predicted 3D poses, computed as geodesic distance \cite{Mahendran2017}.
$L_{trans}$ is the translation loss, defined as the MSE between $\myg_{gt}$ and predicted translation.
$L_{shape}$ is the shape loss, defined as the MSE between $\textbf{dv}_{gt}$ and predicted $\textbf{dv}$.
$L_{uv}$ is the uv-flow loss, defined as the L1 loss between $\textbf{uv}_{gt}$ and the predicted uv-flow. 
Note that the ground truth uv-flows are necessarily incomplete because, from one image, we can only assign a partial texture map.
When generating the digital dataset we also compute visibility masks for the uv-flow, that we exploit for computing the loss only on the visible textels.
$L_{tex}$ is the L1 loss between the $T_{gt}$ and the predicted texture map.
$L_{dt}$ is a texture loss term that encourages the uv-flow to pick colors from the foreground region (see \cite{cmrKanazawa18}).
This term is consistently applied to the full texture. 
Each loss is associated with a weight.
Note that we include both a loss on the model parameters and a loss on silhouettes and projected keypoints. 
Similar to Pavlakos \etal \cite{PavlakosCVPR2018}, we observed that a per-vertex loss improves training compared with naive parameter regression.
The network is implemented in PyTorch.

\input{source/bigfig}

\subsection{Per-instance optimization}
The prediction of the texture map allows us to perform an unsupervised per-instance optimization exploiting the learned feature space and a photometric loss.
Methods that also exploit photometric loss often are assisted by a
segmentation~\cite{Janner2017}, where the loss is computed only in the
foreground region. In this work we do not assume any segmentation
is available at test time, instead we render a 
full image prediction, which requires building a background model~\cite{Schonborn:2015}.
We estimate an average background color exploiting the manual segmentations of
the images used to build the SMALR models for training.
Given an input image, we run the regression network and then perform a per-instance optimization where we keep the network layers fixed and optimize over the feature space variables
(Figure \ref{fig:network} dotted line).
In this way we exploit the correlation between variables learned by the network.
During optimization we minimize the following loss: $L_{opt} =$
\begin{eqnarray}
L_{photo}(I_{input},I) + L_{cam}(\hat{f},f)+L_{trans}(\hat{\myg},\myg),
\end{eqnarray}
where $L_{photo}$ is the photometric loss, computed as the perceptual distance \cite{zhang2018perceptual} between input image and image prediction.
$\hat{f}$ and $\hat{\myg}$ are the initialization values for focal length and translation estimated with the network, respectively.

%% file: source/bigfig.tex
\begin{figure*}
    \centerline{
   \includegraphics[width=0.094\linewidth]{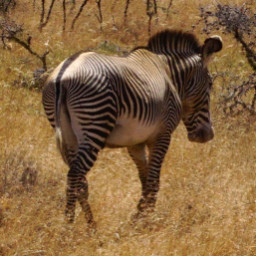} 
   \includegraphics[width=0.094\linewidth]{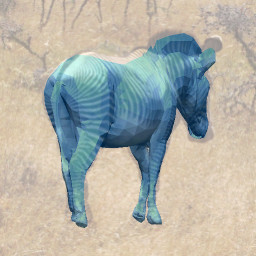} 
   \includegraphics[width=0.094\linewidth]{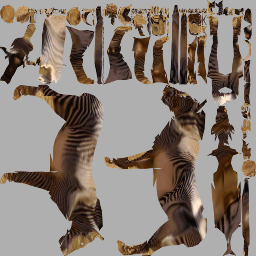} 
   \includegraphics[width=0.094\linewidth]{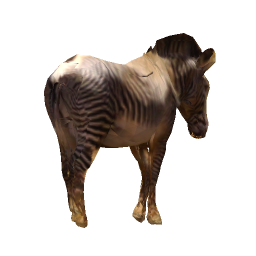} 
   \includegraphics[width=0.094\linewidth]{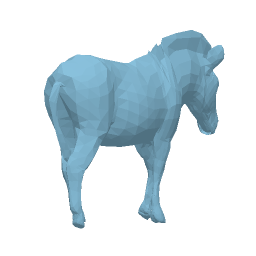} 
 \includegraphics[width=0.094\linewidth]{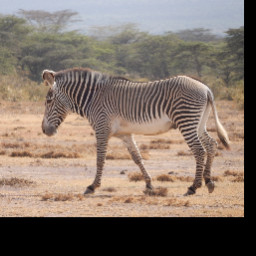} 
   \includegraphics[width=0.094\linewidth]{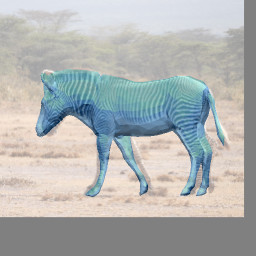} 
   \includegraphics[width=0.094\linewidth]{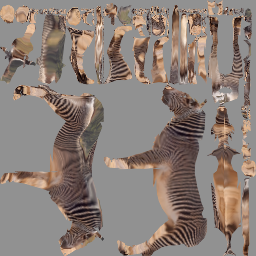} 
   \includegraphics[width=0.094\linewidth]{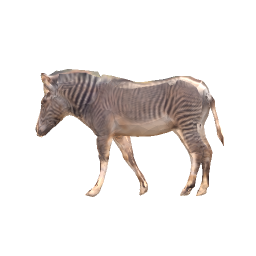} 
      \includegraphics[width=0.094\linewidth]{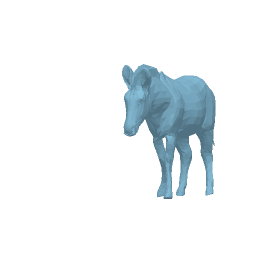} 
      }
    \centerline{
   \includegraphics[width=0.094\linewidth]{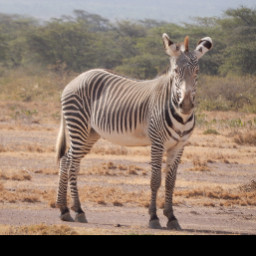} 
   \includegraphics[width=0.094\linewidth]{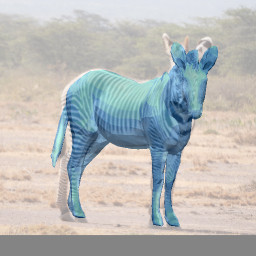} 
   \includegraphics[width=0.094\linewidth]{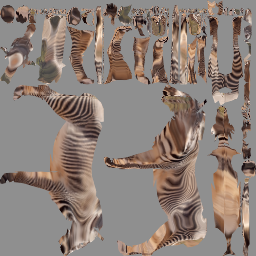} 
   \includegraphics[width=0.094\linewidth]{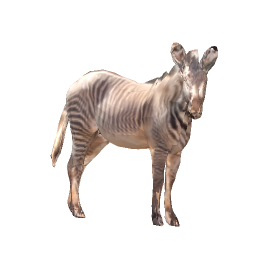} 
   \includegraphics[width=0.094\linewidth]{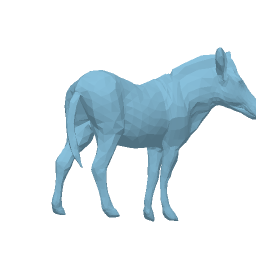} 
 \includegraphics[width=0.094\linewidth]{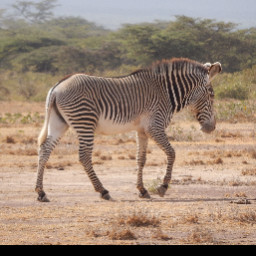} 
   \includegraphics[width=0.094\linewidth]{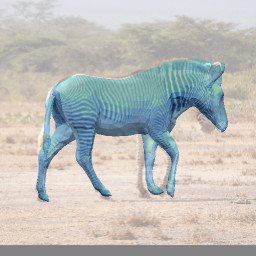} 
   \includegraphics[width=0.094\linewidth]{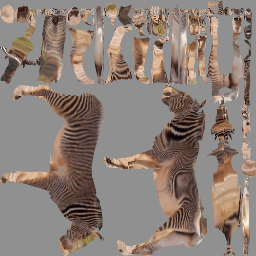} 
   \includegraphics[width=0.094\linewidth]{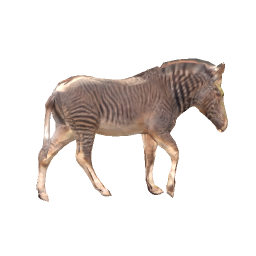} 
      \includegraphics[width=0.094\linewidth]{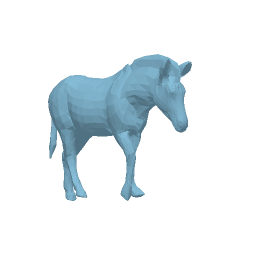} 
    }
    \centerline{
   \includegraphics[width=0.094\linewidth]{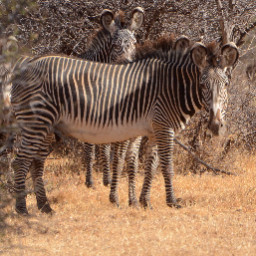} 
   \includegraphics[width=0.094\linewidth]{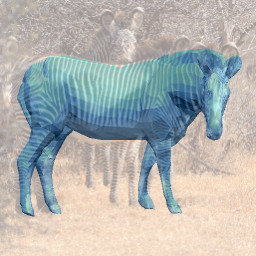} 
   \includegraphics[width=0.094\linewidth]{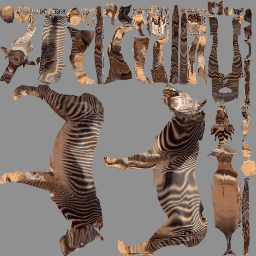} 
   \includegraphics[width=0.094\linewidth]{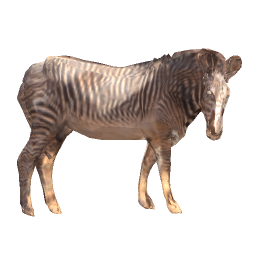} 
   \includegraphics[width=0.094\linewidth]{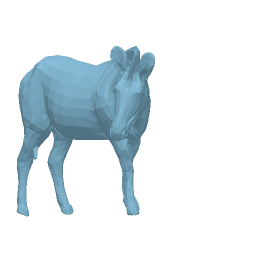} 
 \includegraphics[width=0.094\linewidth]{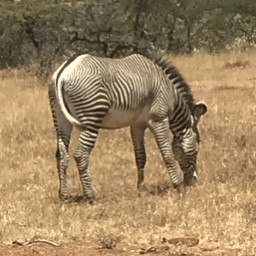} 
   \includegraphics[width=0.094\linewidth]{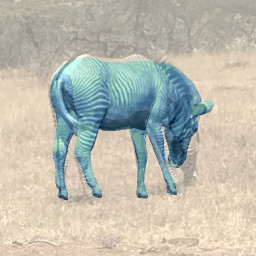} 
   \includegraphics[width=0.094\linewidth]{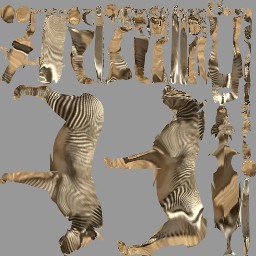} 
   \includegraphics[width=0.094\linewidth]{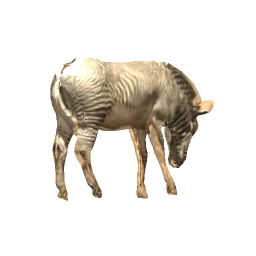} 
      \includegraphics[width=0.094\linewidth]{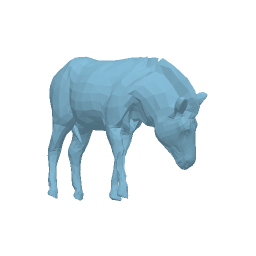} 
    }
   \centerline{
   \includegraphics[width=0.094\linewidth]{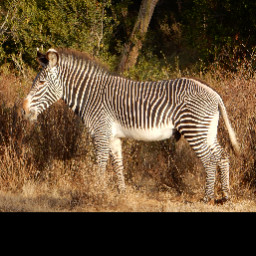} 
   \includegraphics[width=0.094\linewidth]{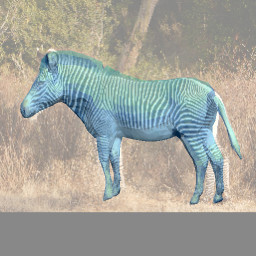} 
   \includegraphics[width=0.094\linewidth]{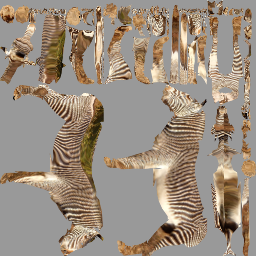} 
   \includegraphics[width=0.094\linewidth]{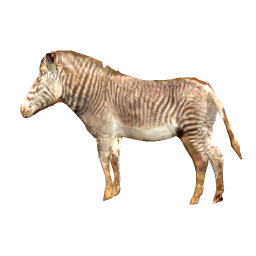} 
   \includegraphics[width=0.094\linewidth]{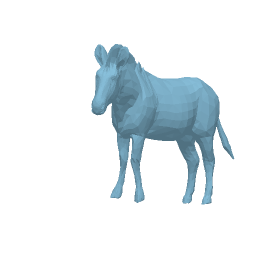} 
 \includegraphics[width=0.094\linewidth]{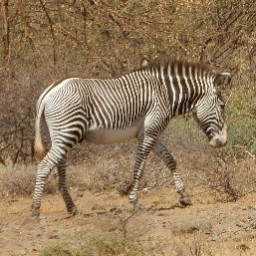} 
   \includegraphics[width=0.094\linewidth]{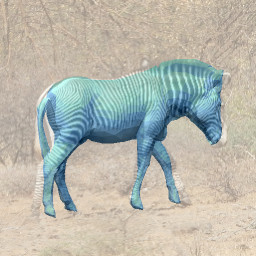} 
   \includegraphics[width=0.094\linewidth]{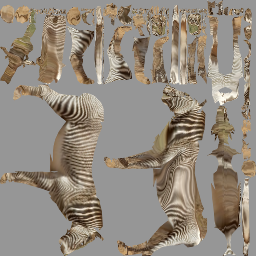} 
   \includegraphics[width=0.094\linewidth]{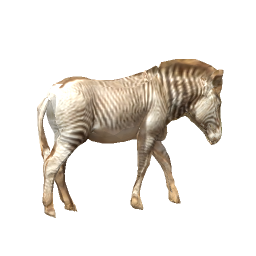} 
      \includegraphics[width=0.094\linewidth]{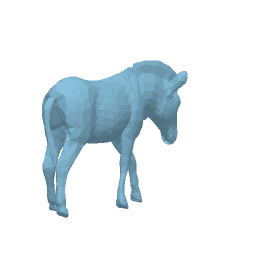} 
    }
   \centerline{
   \includegraphics[width=0.094\linewidth]{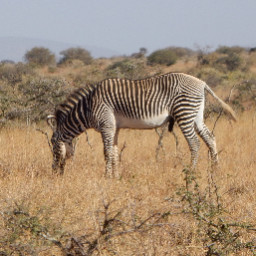} 
   \includegraphics[width=0.094\linewidth]{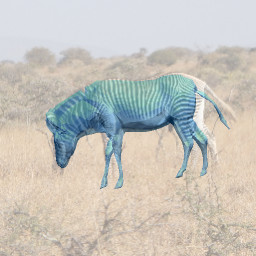} 
   \includegraphics[width=0.094\linewidth]{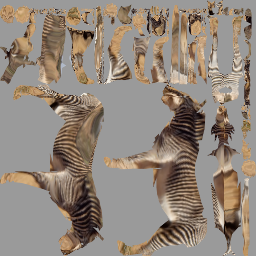} 
   \includegraphics[width=0.094\linewidth]{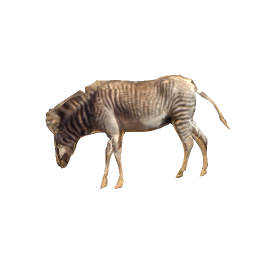} 
   \includegraphics[width=0.094\linewidth]{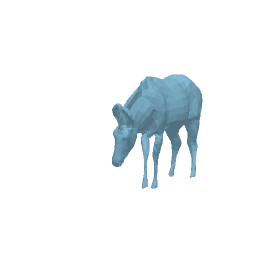} 
 \includegraphics[width=0.094\linewidth]{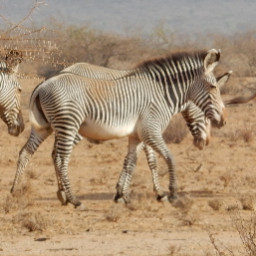} 
   \includegraphics[width=0.094\linewidth]{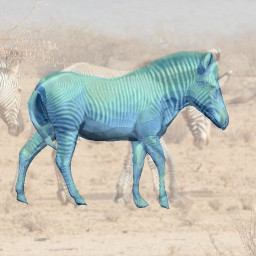} 
   \includegraphics[width=0.094\linewidth]{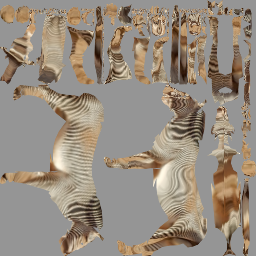} 
   \includegraphics[width=0.094\linewidth]{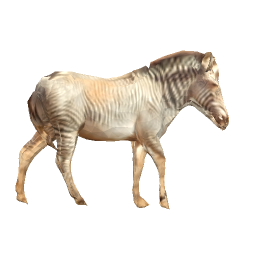} 
      \includegraphics[width=0.094\linewidth]{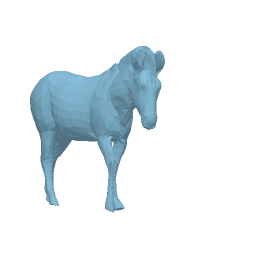} 
    }
  \centerline{
   \includegraphics[width=0.094\linewidth]{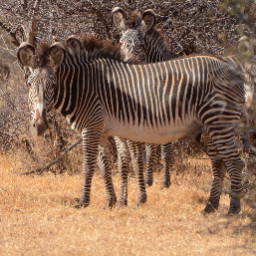} 
   \includegraphics[width=0.094\linewidth]{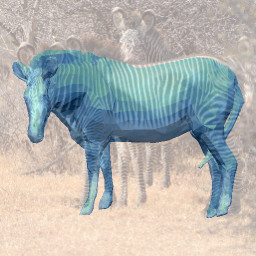} 
   \includegraphics[width=0.094\linewidth]{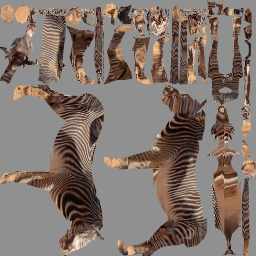} 
   \includegraphics[width=0.094\linewidth]{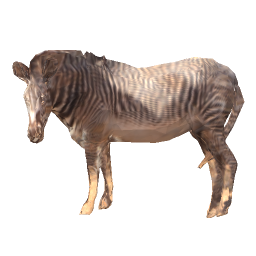} 
   \includegraphics[width=0.094\linewidth]{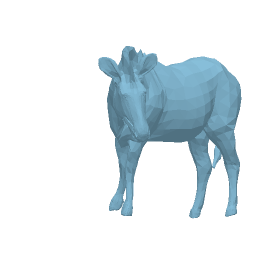} 
 \includegraphics[width=0.094\linewidth]{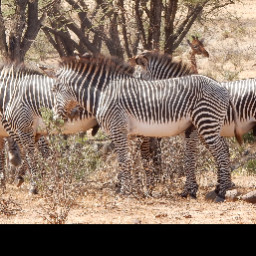} 
   \includegraphics[width=0.094\linewidth]{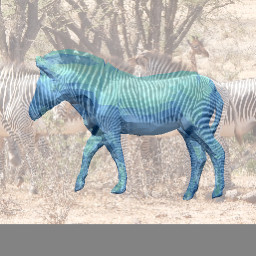} 
   \includegraphics[width=0.094\linewidth]{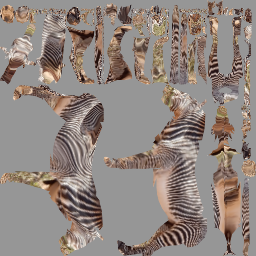} 
   \includegraphics[width=0.094\linewidth]{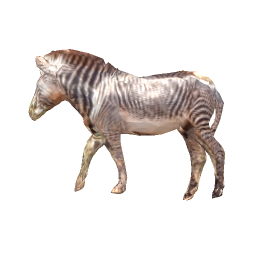} 
      \includegraphics[width=0.094\linewidth]{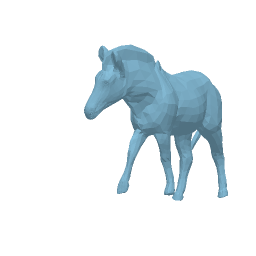} 
    }

   \captionof{figure}{{\bf Results}. On two columns: input image, mesh overlap, predicted texture map, 3D rendering, 3D mesh.
   } 
\label{fig:results}
\end{figure*}

%% file: source/experiments.tex
\section{Experiments}
\noindent\textbf{Direct regression from the image}.
We train two networks: a full network 
and one without the texture prediction module, 
which is therefore trained without the uv-flow loss $L_{uv}$, the image loss $L_{img}$, and the texture map losses $L_{tex}$ and $L_{dt}$.
We train both networks for a budget of 200 epochs, and retain the networks that performed best on the validation set.
We use the Adam optimizer with a learning rate of $0.0001$ and momentum of $0.9$.

We run the prediction network on the set of 200 annotated test images;
see Figure \ref{fig:results} for representative results.
Table \ref{tab:results} reports the average PCK error for the predicted 2D keypoints at two thresholds for the feed-forward prediction (F) and for the network without texture prediction (G).
For comparison with previous work we fit the SMAL model to the testset using the manual segmentations and keypoint annotations (A).
We also evaluate the performance on a synthetic dataset (B).
To illustrate robustness, we perform an experiment adding noise to the input bounding boxes (H).
To quantify the accuracy of the shape estimation, we compute an overlap score as intersection over union 
of the manual image segmentations and predicted animal mask.

\begin{figure}
    \centerline{
   \includegraphics[width=1.0\linewidth]{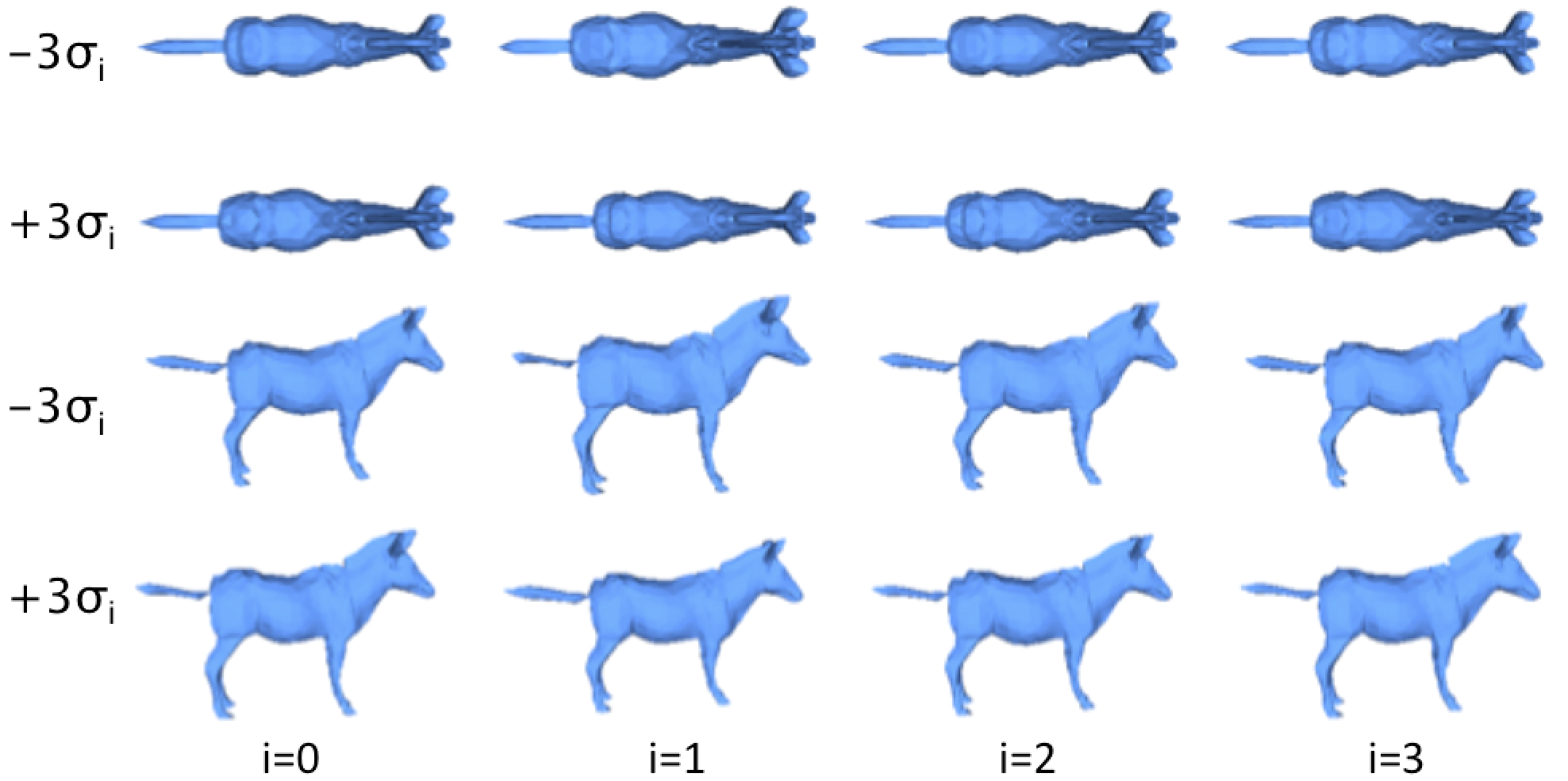} }
 \centerline{
   \includegraphics[width=1.0\columnwidth]{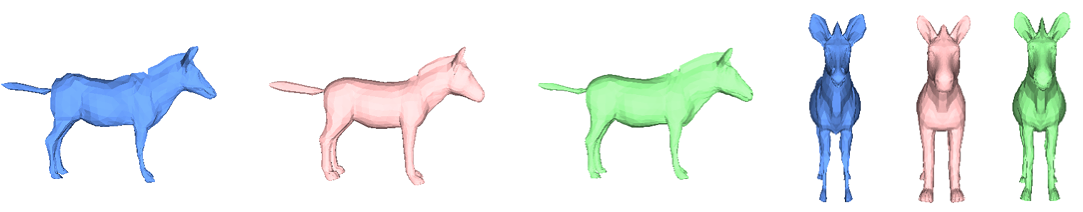}
     }
    \captionof{figure}{{\bf Shape space}. {\em Top}: Visualization of the
      variability of shape in the testset from top and lateral view (see text). We observe variability is in the belly and ears.
    {\em Bottom}: Mean shapes: blue-network, pink-SMAL, green-SMALR.}
\label{fig:shape_space}
\end{figure}

In order to visualize the variability of the estimated shape 
we computed the variance of the $40$ shape features $f_s$ obtained on the test set (see Equation \ref{eq:dv}) and look at the deformations
that are associated with the $4$ features with maximum variance.
In Figure \ref{fig:shape_space} (top) we show $\textbf{v}_{horse}+b-3\sigma_iW_i$ in the odd rows, and
$\textbf{v}_{horse}+b+3\sigma_iW_i$ in the even rows, where $\sigma_i$ is the standard deviation of the 
$i-th$ shape feature and $W_i$ is a row in the learned matrix $W$ (Equation \ref{eq:dv}). 
In order to visualize the difference between the initial SMAL shape space and the shape space learned from the network,
Figure \ref{fig:shape_space} (bottom) shows the mean shape of the network model (blue) obtained by adding the bias $b$ to the SMAL template, the SMAL template (pink) and the average of the SMALR meshes used to create the training set (green). 

\begin{figure}[t]
    \centerline{
   \includegraphics[width=0.21\linewidth]{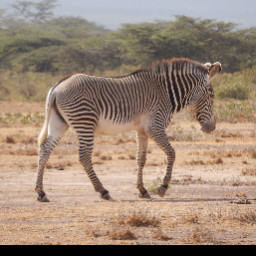} 
   \includegraphics[width=0.21\linewidth]{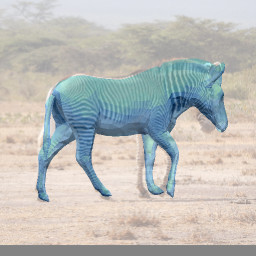} 
   \includegraphics[width=0.21\linewidth]{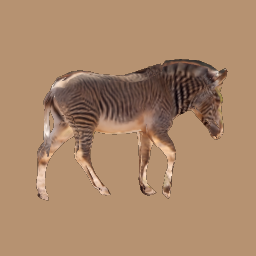} 
   \includegraphics[width=0.21\linewidth]{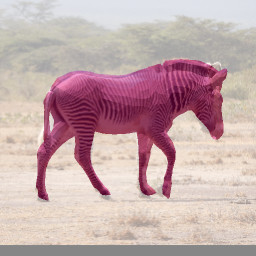} 
      }
    \centerline{
   \includegraphics[width=0.21\linewidth]{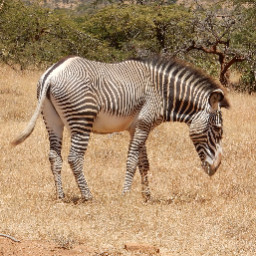} 
   \includegraphics[width=0.21\linewidth]{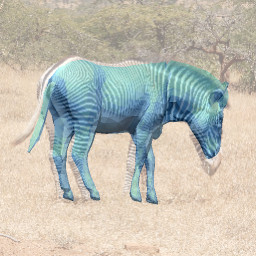} 
   \includegraphics[width=0.21\linewidth]{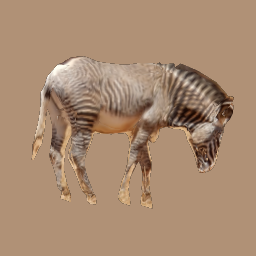} 
   \includegraphics[width=0.21\linewidth]{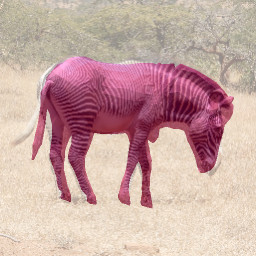} 
      }
    \centerline{
   \includegraphics[width=0.21\linewidth]{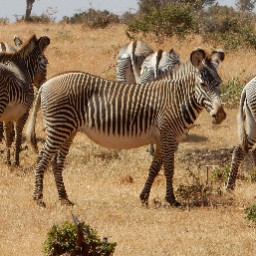} 
   \includegraphics[width=0.21\linewidth]{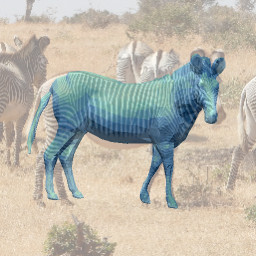} 
   \includegraphics[width=0.21\linewidth]{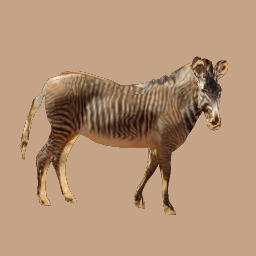} 
   \includegraphics[width=0.21\linewidth]{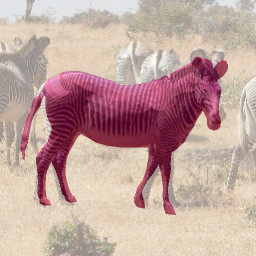} 
      }
    \centerline{
   \includegraphics[width=0.21\linewidth]{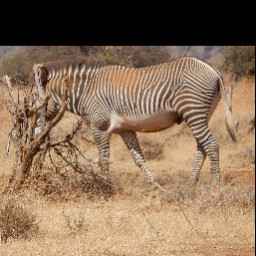} 
   \includegraphics[width=0.21\linewidth]{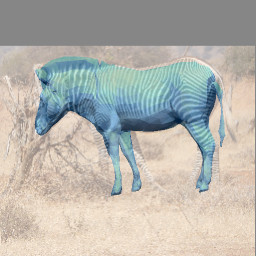} 
   \includegraphics[width=0.21\linewidth]{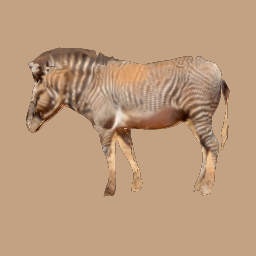} 
   \includegraphics[width=0.21\linewidth]{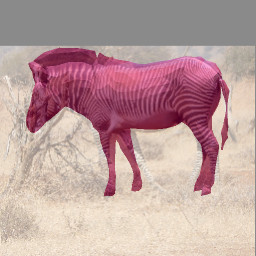} 
      }
   \captionof{figure}{{\bf Per-instance optimization}. Input image, initial network prediction, optimization image and overlap.}
\label{fig:opt}
\end{figure}

\textbf{Per-instance optimization}.
We run per-instance optimization on the whole test set. 
We optimize over a budget of 120 epochs, and retain the solution with lowest photometric loss.
Table \ref{tab:results} shows the performance of the optimization (C). Figure \ref{fig:opt} shows some examples.
For comparison, we also perform per-instance optimization over the model variables (D).

\noindent\resizebox{\columnwidth}{!}{
  \begin{tabular}{||l c c c||} 
 \hline
     Method & PCK@0.05 & PCK@0.1 & IoU \\ 
 \hline\hline
 (A) SMAL (gt kp and seg) & 92.2 & 99.4  &  0.463 \\
 \hline
(B) feed-forward on synthetic &  80.4 &97.1  &  0.423\\
 \hline
(C)  opt features &  {\bf62.3} &{\bf81.6} & {\bf0.422} \\
(D)  opt variables & 59.2& 80.6 & 0.418  \\
 (E) opt features bg img &  59.7& 80.5 & 0.416 \\      
(F) feed-forward pred. &  59.5 &80.3  & 0.416  \\ 
(G) no texture &  52.3 &76.2 &  0.401\\
(H) noise bbox  & 58.7 & 79.9  & 0.415  \\ 
 \hline
 \end{tabular}
 }
\captionof{table}{ {\bf Results.} (A) We compare with SMAL model fitting~\cite{Zuffi:CVPR:2017}, which requires ground
  truth  keypoints and segmentations; (B) We run the network feed-forward prediction on a synthetic dataset; (C) our proposed method;
  (D) per-instance optimization on model variables rather than network features; (F) feed-forward prediction (no optimization);
  (G) feed-forward prediction without texture; (H) feed-forward prediction with noise on the bounding boxes.
}
\label{tab:results}

%% file: source/conclusion.tex
\section{Conclusion}
We have presented the first study of automatic 3D pose, shape and
appearance capture of animals from a single image acquired ``in-the-wild''.
Traditionally the focus of computer vision research has been the human body: dealing with wild animals presents new technical challenges. 
We overcome the lack of training data by creating a digital dataset that combines real appearance with synthesized pose, shape and background.
We remove the need for keypoint detection or segmentations by training an end-to-end network to directly regress 
3D pose, shape, camera, global translation and texture from an image.
We have shown that predicting texture maps helps to recover more accurate pose and shape.
Moreover we have shown that, thanks to the predicted texture map, we can improve results by performing per-instance network-based optimization over the encoder features by exploiting a photometric loss.
In this work we have focused on the Grevy's zebra, but our approach is general and can be applied to other animals or humans.

{\footnotesize {\bf Acknowledgement.} AK is supported by BAIR sponsors. TBW is supported by NSF grant III-1514126.

{\footnotesize {\bf Disclosure.} MJB has received research
gift funds from Intel, Nvidia, Adobe, Facebook, and Amazon.
While MJB is a part-time employee of Amazon, his
research was performed solely at MPI. He is also an investor
in Meshcapde GmbH.}